\documentclass[runningheads]{llncs}
\usepackage[T1]{fontenc}
\usepackage{cite}
\usepackage{graphicx}
\usepackage{comment}
\usepackage{subcaption}
\usepackage{bm}
\usepackage{caption}

\usepackage{algorithm}
\usepackage{algpseudocode}

\usepackage{makecell}
\usepackage{array}
\usepackage{amsmath}
\usepackage{amssymb}
\usepackage{mathtools}
\DeclareMathOperator*{\argmax}{argmax}

\begin{document}
\title{A Bayesian Approach to Clustering via the Proper Bayesian Bootstrap: the Bayesian Bagged Clustering (BBC) algorithm
}
%
%
\author{Federico Maria Quetti\inst{1}
\and
Silvia Figini\inst{2} \and
Elena Ballante\inst{2}}
\authorrunning{F.M. Quetti et al.}
\titlerunning{Bayesian Bagged Clustering algorithm}
%
\institute{Department of Mathematics, University of Pavia, Italy \and
Department of Political and Social sciences, University of Pavia, Italy
}
\maketitle              
\begin{abstract}

The paper presents a novel approach for unsupervised techniques in the field of clustering. A new method is proposed to enhance existing literature models using the proper Bayesian bootstrap to improve results in terms of robustness and interpretability. \\
Our approach is organized in two steps: k-means clustering is used for prior elicitation, then proper Bayesian bootstrap is applied as resampling method in an ensemble clustering approach. Results are analyzed introducing measures of uncertainty based on Shannon entropy. The proposal provides clear indication on the optimal number of clusters, as well as a better representation of the clustered data. Empirical results are provided on simulated data showing the methodological and empirical advances obtained.
\\

\keywords{Proper Bayesian Bootstrap  \and Bayesian Clustering \and Bagging \and Fuzzy Clustering 
}
\end{abstract}

\section{Introduction}

Cluster analysis is the field of Machine Learning (ML) that deals with partitioning data by finding groups of similar units in an unsupervised framework. Clustering techniques are widely used in various applications, where capturing the inherent structure of data plays a pivotal role for the analysis. \\
A large body of literature exists in the field regarding methods of clustering \cite{jain1999data, Elements2009, Jaeger2023}; yet, being the problem unsupervised, the research on improvements to existing methods is still an open point, leaving room to further developments. In particular, the integration of ML with techniques from Bayesian statistical learning has been shown to provide significant improvements in the supervised framework by \cite{galvani2021} and \cite{ballante2023extension}. 
In an unsupervised setting, it has been shown in the literature that the application of bagging techniques to standard clustering methods improves results and brings new information of fuzzy clustering type \cite{dudoit2003bagging}. \\
The aim of this work is to extend the current state of the art in clustering, adopting Bayesian Bootstrap techniques in unsupervised learning resorting to a prior knowledge integration scheme. 
The rest of the paper is structured as follows: Section 2 reviews the existing literature on clustering, bootstrap, and proper Bayesian bootstrap. Section 3 introduces our proposal and Section 4 reports empirical evidence. Conclusions are drawn in Section 5. 

\section{Background}
In this paper a traditional clustering problem is faced integrating a classical partition based clustering algorithm with a Bayesian bootstrap method. In the present section, a review of the background topics is described, and particular emphasis is given to the Bayesian non-parametric approach to ensemble learning. 

\subsection{Bootstrap}

\subsubsection{Bootstrap methods}
Bootstrap is a statistical resampling technique used to estimate the distribution of a statistic, by providing an approximation of the empirical distribution function of data. Formally, given $\{X_1,\dots,X_n\}$ i.i.d. realization of a random variable $X$, we are interested in estimating the distribution of a functional $\Phi(F,X)$, depending on the cumulative distribution function $F$ of the variable $X$. In order to generate the distribution of the estimator $\hat{\Phi}$ for the functional, an approximation $F^*$ for the cumulative distribution of $X$ is 
needed. \\
A first method was proposed by Efron in \cite{efron1992bootstrap}:
the approximation for the cdf is obtained by generating replications with replacement from the sample.
The procedure consists in 
drawing a weights vector for the observations from a Multinomial distribution, and defining the estimate for the  population cdf as:
\begin{equation}
    F^*(x)= \sum_{i=1}^n \frac{w_i}{n}  \mathbb{I}[X_i \leq x]
\end{equation}
with $(w_1, \dots, w_n) \sim \textit{Mult}(n, \frac{1}{n}\mathbf{1}_n)$ \ and \  $\mathbb{I}[X_i \leq x]$ indicator function. 
\\ 
An alternative method was proposed by Rubin in \cite{rubin1981bayesian},  called Bayesian bootstrap: the method is similar to Efron's but modifies the definition of the weights, which are obtained by sampling from a Dirichlet distribution. The discrete cdf approximating the empirical is then:
\begin{equation}
F^*(x) = \sum_{i=1}^{n} w_i \mathbb{I}[X_i \leq x]\end{equation}
where $(w_1, \dots, w_n) \sim D(\mathbf{1}_n)$. \\
The two bootstrap methods are asymptotically equivalent \cite{Lo1987} and first order equivalent from the predictive point of view, as the conditional probability of a new observation is estimated by only using observed values from the sample \cite{galvani2021}. 

\subsubsection{Proper Bayesian Bootstrap}
In order to present the bootstrap method central to the following work, a brief digression on the theoretical foundations of Bayesian non-parametric learning is needed. The framework presented here refers to the univariate case. \\
Given an exchangeable sequence of real random variables $\{X_n\}$ on a probability space $(\Omega,\mathcal{F}, P)$, De Finetti's Representation Theorem ensures the existence of a random distribution $F$ conditionally on which the variables are i.i.d with distribution $F$. Let $\Phi(F,X)$ be a functional depending on the random distribution and on the sample values $X$.
The Bayesian approach to the evaluation of the conditional probability requires to elicit a prior distribution for $F$ on the space of distribution functions, with the aim of using the posterior of $F$ to estimate the distribution of the functional $\Phi$ given the sample values. 
Ferguson, in a fundamental paper on Bayesian approach to non-parametric statistics \cite{ferguson1973bayesian}, defined a prior for the random distribution, referred to as Dirichlet process.
Given a proper distribution function $F_0$ interpreted as the
prior guess at $F$, and a positive real number $k$ interpreted as a confidence parameter in this guess, $kF_0$ is the parameter of the process denoted as $\mathcal{DP}(kF_0)$.
The relevance of this process definition is motivated by the fact that it is conjugate: given a random sample $\{x_1, \dots, x_n\}$ from $F \sim \mathcal{DP}(kF_0)$, the posterior is again a Dirichlet process, with updated parameter.
\[
F | X \sim \mathcal{DP}((k + n)G_n) \tag{1}
\]
where
\[
G_n = \frac{k}{k + n}F_0 + \frac{n}{k + n}F_n \tag{2}
\]
The parameter of the Dirichlet process, given the data, becomes a convex combination of the prior guess $F_0$ and the empirical cdf $F_n$. Posterior estimations of different functionals $\Phi(F,X)$ are then easily computable by first updating the parameters of the prior. \\
The first approach to prior knowledge integration to the process of resampling data was proposed by Muliere and Secchi in \cite{muliere1996bayesian}: a proper distribution function $F_0$ is introduced as the baseline parameter, and the sampling is performed from the posterior which is the process with baseline $(k+n)G_n$. 
If $k \rightarrow 0$ then a non-informative prior on $F_0$ is considered, and we fall back on the proposal of Rubin; if $k \rightarrow \infty$ the parameter of the posterior Dirichlet Process is reduced to $F_0$, so that the empirical information is of no relevance in resampling.
The proposal is detailed in Algorithm 1.

\begin{algorithm}[H]
\caption{Proper Bayesian bootstrap}
\begin{algorithmic}[1]
\Procedure{ProperBayesianBootstrap}{$B, m, k, F_0, L = \{x_1,\ldots, x_n\}$}
\For{$b$ in $1:B$}
    \State Generate $m$ observations $x_1^*, \ldots, x_m^*$ from $(k + n)^{-1} (k F_0 + n F_n)$.
    \State Draw $w_1^b, \ldots, w_m^b$ from $D(\frac{n+k}{m}, \ldots, \frac{n+k}{m})$, weights for the observations. 
\EndFor
\EndProcedure
\end{algorithmic}
\end{algorithm}


\subsection{Clustering}
Clustering is the process of grouping data points together based on their similarity with respect to certain features, aiming to uncover inherent patterns or structures within the dataset. Literature methods \cite{jain1999data} differ on the approach chosen and with respect to the similarity definition used, the most common ones are:
\begin{itemize}
\item hierarchical clustering. It involves organizing data into a tree-like structure, where clusters are nested within one another according to a similarity measure \cite{hierarchical};\\
\item density-based clustering. It assigns groups based on regions of high density separated by regions of low density \cite{dbscan};\\
\item partitional clustering. Given the number of clusters, creates a partition of data points iteratively, aiming to minimize a cost function \cite{Elements2009}. \\
\end{itemize}
In this paper, the focus is on K-means algorithm \cite{jmac1967} as representative of partition-based clustering methods. 
Chosen a number of clusters $K$, the cluster centroids are picked from the data at random, and the partition is defined so that each point is assigned to the cluster with nearest centroid $\boldsymbol{\mu}_k$ in $L^2$ norm. The centroids are then updated as means of points in each component, and the assignment repeated. At each step, this procedure amounts to minimizing the total within sum of squares of the $K$ clusters:
\begin{equation}
    WSS = \sum_{k=1}^{K} \sum_{i \in C_k} \left\| \mathbf{x}_i - \boldsymbol{\mu}_k \right\|^2
\end{equation}
Typically, the procedure quickly converges to a local optimum. \\
Another type of clustering paradigm, called fuzzy clustering, differs from the above as labels are not assigned as one-hot vectors. Instead, memberships $u_k$ are assigned to data points and indicate the degree to which they belong to each cluster: $u_k \in [0,1], \ \ \ \sum_{k=1}^K u_k = 1$. 
For fuzzy c-means \cite{fuzzycmeans}, the counterpart of standard K-means, at each iteration the cost function to optimize given the centroids $\boldsymbol{c}_k$ is $J_m$, $m$ being an hyperparameter:
\begin{equation}
\begin{split}
    & \boldsymbol{c}_k = \frac{\sum_{i=1}^{N} u_{ik}^m \mathbf{x}_i}{\sum_{i=1}^{N} u_{ik}^m} \\
    & J_m = \sum_{k=1}^{K} \sum_{i=1}^{N} u_{ik}^m \|\mathbf{x}_i - \boldsymbol{c}_k\|^2
\end{split}
\end{equation}
The main difference from the classical K-means is that the output is not limited to an hard assignment, but the memberships returned give an idea of the confidence in the possible assignment. Note that in the following, differently from fuzzy c-means, memberships are not given by an optimization procedure. Instead, they are the result of the aggregation of labels obtained applying clustering on the bootstrap replicas.

\subsubsection{Bootstrap in clustering}
In this work, a bagging procedure where results obtained for each bootstrap replica are aggregated is applied. The approach stems from the work of \cite{dudoit2003bagging}: the methodology linking bootstrap with clustering is drawn from the algorithm BagClust1, that we outline.
Given a learning set $L$, the number of bootstrap samples $B$, the chosen number of clusters $K$, and a 
clustering method $P$, the algorithm applies the clustering procedure $P$ to the original learning set $L$ obtaining initial cluster labels. Afterwards, for each bootstrap sample, it applies the clustering procedure again and permutes the cluster labels to maximize overlap with the original clustering. After $B$ iterations, it assigns an aggregated cluster label to each observation based on majority vote. Moreover, the algorithm retrieves a fuzzy type of result by also recording the cluster memberships, giving information on the confidence of the resulting label. \\

\begin{algorithm}[H]
\caption{BagClust1}
\label{bagged_clustering}
\begin{algorithmic}[1]
\Procedure{BagClust1}{$L = \{x_1,\ldots, x_n\}, B, K, P$}
    \State Apply the clustering $P$ to the learning set $L$ obtaining cluster labels:
    \State \hspace{\algorithmicindent} $P(x_i; L) = \hat{y}_i$ for each observation $x_i$, $i = 1, \ldots, n$, $\hat{y}_i, i = 1, \ldots, K$.
    \For{$b = 1$ to $B$}
        \State Form the $b$-th bootstrap sample $L_b = (x_{b1}, \ldots, x_{bn})$.
        \State Apply the procedure $P$ to the bootstrap replica $L_b$ obtaining cluster labels $P(x_{bi}; L_b)$ for each observation in $L_b$.
        \State Permute the cluster labels assigned to the bootstrap learning set $L_b$ for maximum overlap with the original clustering labels: 
        let $S_K$ denote the set of all permutations of the integers $1, \ldots, K$. Find $\tau_b \in S_K$ that maximizes:  $\sum_{i=1}^n I_{\tau_b}(P(x_{bi}; L_b)) = P(x_{bi}; L)$.
        \hspace{\algorithmicindent}
    \EndFor
    \State For the data points, record cluster memberships as the proportion of votes in favor of each cluster assignment: $u_k(x_i) = \frac{\sum_{\{b: x_i \in L_b\}} I_{\tau_b}(P(x_i; L_b))=k}{|\{b: x_i \in L_b\}|}$. 
    Assign a bagged cluster label for each observation $i$ by majority vote:
$\text{argmax}_{1 \leq k \leq K} u_k(x_i)$.
    
\EndProcedure
\end{algorithmic}
\end{algorithm}

\section{Our methodological proposal: the Bayesian Bagged Clustering (BBC)}

The proposal of this work is twofold. \\ In the first part, the BBC clustering algorithm is proposed, aimed at bettering the chosen algorithm 
with improved stability as well as additional information about the uncertainty in the assignments for the dataset. \\
Directly from the results of the above approach, the second part of the proposal focuses on its exploitation by discussing an optimal choice scheme for the number of clusters $K$ detected in the dataset.

\subsection{Clustering procedure}

The BBC procedure is divided in two parts: firstly, cluster information about data is retrieved, from which the prior $F_0$ is defined; secondly, proper Bayesian bootstrap is performed to find clustering results. \\
In the initial step, we apply the partitioning $P$ with a chosen number of clusters $K$ on data. 
This information is used to define a suitable baseline prior for the generating process underlying bootstrap resampling. 
$F_0$ is imposed as the cumulative of a suitable Gaussian mixture probability density, as follows:
\begin{equation}   f_{\boldsymbol{\theta}} = \sum_{j=1}^K p_j f_{\boldsymbol{\mu_j, \Sigma_j}}
\end{equation}
where $\boldsymbol{\theta} = (p_j, \boldsymbol{\mu_j},\boldsymbol{\Sigma_j}), \ j=1,\dots, K$ are the mixture parameters, and $f_{\boldsymbol{\mu_j, \Sigma_j}} \sim \mathcal{N}(\boldsymbol{\mu_j, \Sigma_j})$ is the multivariate Gaussian distribution with mean $\boldsymbol{\mu_j}$ and covariance matrix $\boldsymbol{\Sigma_j}$. \\ The shape of the prior $F_0$ is determined via the parameters of the distribution calculated from the first step: the mixture parameters are associated to each component $j$.
The weight $p_j,\ 0 \leq p_j \leq 1, \ \sum_{j=1}^K p_j = 1$, is evaluated as the proportion of data assigned to cluster $j$ in the dataset initial cluster labeling. The mean $\boldsymbol{\mu_j}$, representing the cluster centroid, is taken as centroid $j$ got from the partitioning $P$.
The variance matrix $\boldsymbol{\Sigma_j}$ is linked to the informativeness of the prior. In the analysis described in Section \ref{sec:results} this is defined as the empirical covariance matrix of the data points in cluster $j$, $\boldsymbol{\Sigma^*_j}$, multiplied for a constant value $s$ as to consider different concentrations.
Upon selection of a value for $s$ in the variance matrix  $\boldsymbol{\Sigma_j} = s\boldsymbol{\Sigma^*_j}$, a prior pdf for the whole dataset is entirely defined. \\
The second step of the procedure is based on proper Bayesian bootstrap: $m=n$ observations are generated from the convex combination of the defined prior and empirical cdf defined as:
\begin{equation*}
    G_n = (k+n)^{-1}(k F_0 + nF_n)
\end{equation*}
where $k$ is the assigned confidence parameter.
The proper Bayesian bootstrap resamples present newly sampled values as well as original dataset values, which are the focus of the cluster labels assignment.
K-means is applied to the $B$ resampled learning sets, obtaining a cluster partitioning for each. As a result, each of the original data points will be assigned to a given cluster a certain total number of times: cluster memberships are evaluated as the fraction between this total and the number of times the point has been selected overall. \\
The procedure finally gives an aggregated value of the cluster label for the original data points from the memberships, as $\argmax_{1 \leq k \leq K} u_k(\boldsymbol{x_i})$.







\begin{algorithm}
\caption{Proposal: BBC clustering procedure}
\label{proposal}
\begin{algorithmic}[1]
\Procedure{Bayesian Bagged Clustering }{$L=\{x_1,\ldots, x_n\}, K, P, m, k, s$}
    \State Apply $P$ to the learning set $L$, retrieving the cluster labels for the data points.
    \For{$j=1$ to $K$}
        \State Evaluate the parameters of component $j$: $p_j= n_j /n$, $n_j$ being the number of points in cluster $j$; $\boldsymbol{\mu_j}$ as the $j$th centroid; $\boldsymbol{\Sigma_j} = s\boldsymbol{\Sigma^*_j}$ from the empirical covariance matrix of points in cluster $j$ weighted with the parameter $s$. 
    \EndFor
    \State Define the prior pdf: $ f_{\boldsymbol{\theta}} = \sum_{j=1}^K p_j  \mathcal{N}(\boldsymbol{\mu_j},\boldsymbol{ \Sigma_j})$.
    \For{$b=1$ to $B$}
        \State Generate $n$ observations from $(k+n)^{-1}(k F_0 + n F_n)$
        \State Draw $w_1^b, \dots, w_n^b$, from $ D(\frac{k+n}{n}, \dots, \frac{k+n}{n})$ weights distribution of the observations.
        \State Perform $P$ on the dataset obtained above, assigning labels. 
        \State Permute the cluster labels assigned to the data points that come from the original dataset for maximum overlap with the original clustering labels.
        \State Record the labels of the data points which come from the original dataset.
    \EndFor 
    \State For the data points, update the cluster memberships $u_k(x_i)$, defined as the proportions of votes from the last step. \State Assign final cluster labels for each data point equal to $k$ corresponding to the highest cluster membership: $\text{argmax}_{1 \leq k \leq K} u_k(x_i)$.
\EndProcedure
\end{algorithmic}
\end{algorithm}


\subsubsection{Geometrical interpretation}

\begin{figure}[h]
    \centering
\includegraphics[width=0.75\linewidth]{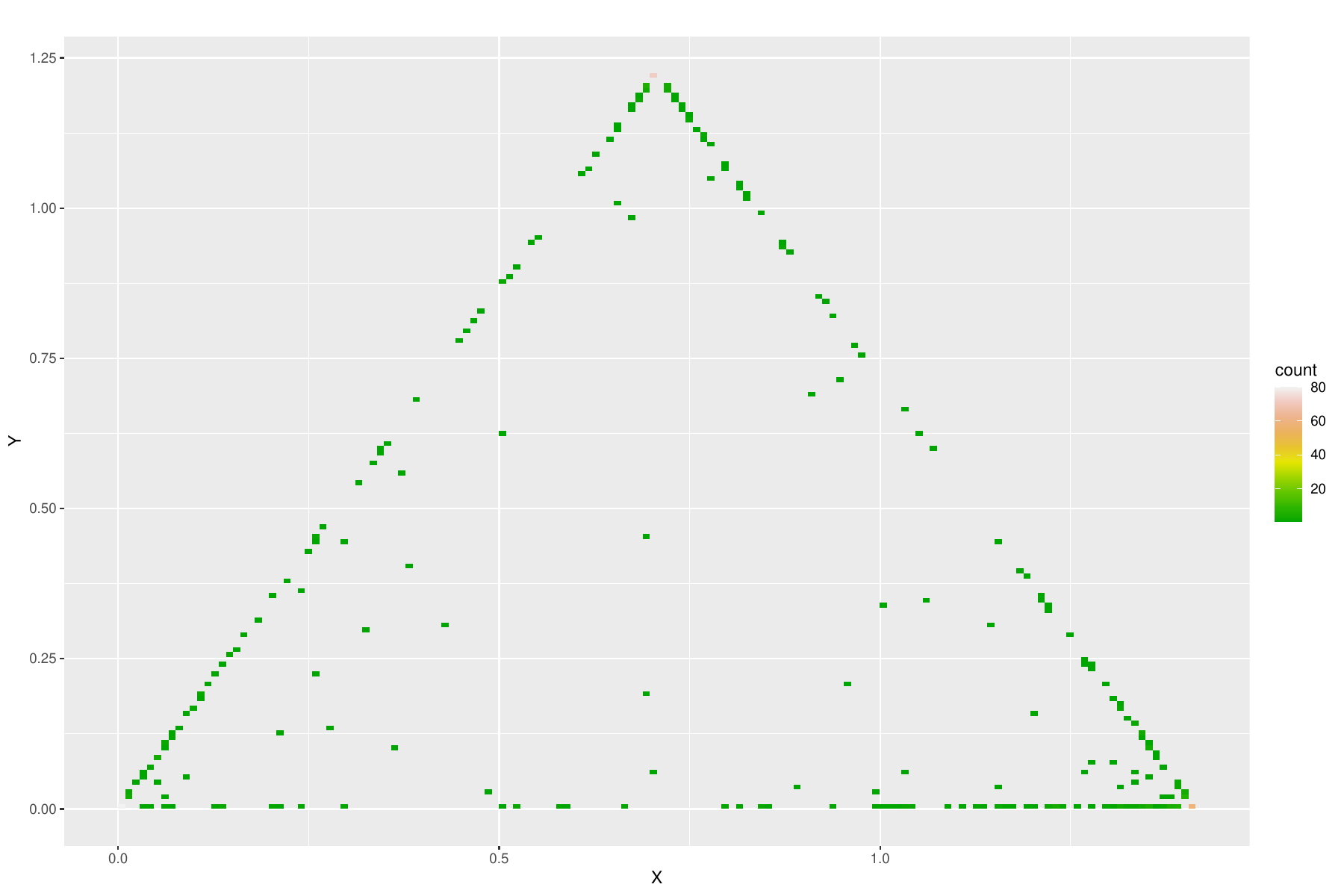}
    \caption{Simplex representation of membership vectors obtained with the proper Bayesian bootstrap, choosing $K=3$. The dataset is comprised by 500 points in 2 dimensions, uniformly generated.}
    \label{simplex}
\end{figure}

As each vector $\boldsymbol{u_k}(\boldsymbol{x_i})$ describes a probability distribution over the cluster labels, a representation of the original data space of points in a $K-1$ dimensional simplex is induced from the definition. It follows that the concentration of points in the simplex will depend on the concentration of the membership assignment to few components, hence on the optimality of assignment.  The ideal scenario would be of points close to the vertices of the simplex; a real scenario is shown in Fig. \ref{simplex}, for the case of a uniform dataset in 2 dimensions while clustering with $K=3$. It can be seen that points in the worst case tend to be assigned with uncertainty to more than one cluster.

\subsection{Optimal choice of $K$}

From the first part of the proposal, we recover the $K$-dimensional vector of cluster memberships of data points $\boldsymbol{x_i}$, denoted as $ \boldsymbol{u}(\boldsymbol{x_i})$, $i = 1, \dots, n$. 
The vectors evaluated on the dataset will depend on the parameters of the proper Bayesian bootstrap, on the chosen clustering algorithm $P$ and in particular on the number of clusters $K$. This fact motivates the analysis of their behaviour under clustering algorithm $P$, with fixed parameters, for varying $K$, in order to recover the underlying cluster structure of the dataset. In the clustering procedure via proper Bayesian bootstrap, $K$ is the number of components of the prior; moreover, each component parameter (weight, mean, variance) depends on $K$. Under clustering, each replica by definition of the generating process is expected to exhibit an intermediate behaviour between the original dataset and the prior model.
Therefore, the algorithm is effectively enforcing a $K$-cluster structure of data by implementing the prior as prescribed.  \\
The fundamental idea of our proposal is that better choices of $K$ lead to easily assignable labels for the dataset, because the algorithm is more able to disambiguate between clusters.\\
%
To quantify the uncertainty about the membership assignments, we seek to determine how the weight of the components is distributed over the normalized membership vector $\boldsymbol{u}(\boldsymbol{x_i})$. To do so, we employ the following two measures:  
\begin{itemize}
    \item     $S(\boldsymbol{u}(\boldsymbol{x_i})) = - \sum_{i=1}^k u_k(\boldsymbol{x_i}) \log_2 u_k(\boldsymbol{x_i})$, Shannon entropy of the vector, quantifies how the decision is dispersed between every vector component
    \item 
    $S_{l,m}(\boldsymbol{u}(\boldsymbol{x_i})) = - (\frac{u_l}{u_l+u_m} \log_2 \frac{u_l}{u_l+u_m} + \frac{u_m}{u_l+u_m} \log_2 \frac{u_l}{u_m+u_m})$, defined as the Shannon entropy of the normalized two component vector $(u_l,u_m)$, quantifies the pairwise indecision between clusters $l,m$ in labeling the data point.
\end{itemize}
The proposed measures lead to the following observations about the expected results. If the number is optimal, one expects the results of the algorithm to give the most crisp assignments of data points to the clusters: from an information theory viewpoint, the smallest mean value of $S$ as function of $K$ corresponds to the best choice of number of clusters. \\ Moreover, for each $K$, the arguments ${l,m}$ of the maximum assumed by the dataset average of $S_{l,m}$ indicate which two clusters are most ill defined as separated instead of joint; the corresponding value of the measure quantifies the worst case of pairwise indecision stemming from the choice of $K$.  
\\
The implementation of this line of reasoning is as follows: the proposed clustering procedure is performed with multiple values of $K$ in a range of plausible cluster numbers; furthermore, for each case we repeat the procedure with different choices of prior corresponding to different values of $s$.
While for clustering the selection of specific parameters is required, for this part of the proposal the usage of different values of the parameter $s$ ensures coverage of multiple cases with the aim of robustly assessing the behaviour of the measures under every $K$: different scenarios of informativeness of the prior correspond to different parameter choices, as the cluster structure modeled by the prior becomes increasingly overlapped with increasing $s$. The confidence parameter $\omega = \frac{k}{k+n}$ is set to $0.5$ as to equally weigh prior and empirical distribution.
\\ One retrieves the optimal number of clusters as indicated by the two measures: for the first one, the value of $K$ corresponding to the minimum of $\Bar{S} = \sum_{i=1}^N S(\boldsymbol{u}(\boldsymbol{x_i}))/N$; for the second one, the value of $K$ corresponding to the minimum of $\Bar{S}_{l,m} = \mathrm{max}_{(l,m)} \sum_{i=1}^N S_{l,m}(\boldsymbol{u}(\boldsymbol{x_i}))/N$. \\
Finally, the joint usage of the two parameters leads to a taxonomy of clustering results for the original dataset based on information theory, aimed at enriching the understanding about the behaviour for different choices of $K$.

\section{Results}\label{sec:results}

In this section we present results on the clustering proposal as well as examples of analysis of the optimal number of cluster of various datasets performed using the proposed method. 

\subsection{Clustering proposal results}

In order to show how our proposed method works, results obtained for the clustering algorithm applied on the Iris dataset are shown. The results are presented as contingency tables where rows represent true clustering labels while columns the predicted ones. In this part the number of clusters $K$ is considered a known parameter.\\ 
The Iris dataset, extensively studied and considered as a benchmark for clustering applications, is comprised of 150 data points under 4 features; it comes with the true cluster labels of the points, to be tested against the assignment of the clustering procedure. \\
Against the benchmark results for K-means, shown in Table \ref{tab:1}, our proposal is tested with different choices of parameter; results for different values of the variance parameter $s$, and of the confidence $w$, are shown in Table \ref{tab:s} and Table \ref{tab:w}, respectively. \\
From Table \ref{tab:w}, we see that the method is robust when the prior has variance $s=1$, meaning that the model imposed for the dataset probability distribution function (pdf) is well posed. The results on the sensitivity analysis on the variance described in Table \ref{tab:s} show that increasing $s$ the cluster structure is not captured well, worsening the performance in cluster 2. These results confirm the necessity of an accurate tuning of model parameters. 

\begin{table}
\centering
    \begin{tabular}{r|rrr}
      \hline
      & 1 & 2 & 3 \\ 
      \hline
      1 & 50 & 0 & 0 \\ 
      2 & 0 & 48 & 2 \\ 
      3 & 0 & 14 & 36 \\ 
      
    \end{tabular}
    \vspace{2ex}
    \caption{Clustering results for classical K-means clustering.}
    \label{tab:1}
\end{table}

\begin{table}
\centering
\begin{minipage}{0.45\textwidth}
\centering
\begin{tabular}{r|rrr|rr}
  \hline
 & 1 & 2 & 3 & $s$ & $w$ \\ 
  \hline
1 & 50 & 0 & 0 & 1 & 0.1 \\ 
  2 & 0 & 47 & 3 & 1 & 0.1 \\ 
  3 & 0 & 13 & 37 & 1 & 0.1 \\ 
  \hline
  1 & 50 & 0 & 0 & 10 & 0.1 \\ 
  2 & 1 & 46 & 3 & 10 & 0.1 \\ 
  3 & 0 & 15 & 35 & 10 & 0.1 \\ 
  \hline
  1 & 50 & 0 & 0 & 100 & 0.1 \\ 
  2 & 1 & 2 & 47 & 100 & 0.1 \\ 
  3 & 0 & 0 & 50 & 100 & 0.1 \\ 
\end{tabular}
\vspace{2ex}
\subcaption{Contingency tables for different choices of variance parameter $s$.}
\label{tab:s}
\end{minipage}
\hfill
\begin{minipage}{0.45\textwidth}
\centering
\begin{tabular}{r|rrr|rr}
  \hline
 & 1 & 2 & 3 & $s$ & $w$ \\ 
  \hline 
  1 & 50 & 0 & 0 & 1 & 0.3 \\ 
  2 & 0 & 46 & 4 & 1 & 0.3 \\ 
  3 & 0 & 13 & 37 & 1 & 0.3 \\ 
  \hline
  1 & 50 & 0 & 0 & 1 & 0.5 \\ 
  2 & 0 & 45 & 5 & 1 & 0.5 \\ 
  3 & 0 & 10 & 40 & 1 & 0.5 \\ 
  \hline
  1 & 50 & 0 & 0 & 1 & 0.7 \\ 
  2 & 0 & 45 & 5 & 1 & 0.7 \\ 
  3 & 0 & 9 & 41 & 1 & 0.7 \\ 
\end{tabular}
\vspace{2ex}
\subcaption{Contingency tables for different choices of confidence $w$.}
\label{tab:w}
\end{minipage}
\vspace{2ex}
\caption{Results for the proposed method varying the parameters related to the variance of the prior $s$ and the confidence given to the prior $w$.}
\label{Tab2}
\end{table}

\subsection{Optimal choice of $K$ results}

In order to showcase the interpretation stemming from the proposal, evaluations on the synthetic datasets described in Table \ref{table:datasets} are shown in the following. For the case of true number of clusters $K=3$ and dimension of the points $p=2$, the datasets are generated differing by specific ground truth characteristics: overlaps between cluster components, different numerosity between cluster components, different covariances between features i.e. different shapes.

\begin{table}[H]
    \centering
    \begin{tabular}{|c|c|c|c|c|c|}
        \hline
        Dataset & Dimension $p$ & \makecell{Number of \\ Clusters $K$} & \makecell{Points in \\ each component} & Centroids & Covariance $\Sigma$ \\ \hline
       1& 2 & 3 & 33, 33, 33 & $\begin{pmatrix} (1.5, 0) \\ (-1.5,0) \\ (0, \frac{3\sqrt{3}}{2}) \end{pmatrix}$ & $\begin{pmatrix} 1 & 0  \\ 0 & 1 \end{pmatrix}$ \\
        \hline
        2&2 & 3 & 99, 66, 33 & $\begin{pmatrix} (1.5, 0) \\ (-1.5,0) \\ (0, \frac{3\sqrt{3}}{2}) \end{pmatrix}$ & $\begin{pmatrix} 1 & 0  \\ 0 & 1 \end{pmatrix}$ \\
        \hline
        3&2 & 3 & 33, 33, 33 & $\begin{pmatrix} (1, 0) \\ (-1,0) \\ (0, \sqrt{3}) \end{pmatrix}$ & $\begin{pmatrix} 1 & 0  \\ 0 & 1 \end{pmatrix}$ \\ \hline
        4&2 & 3 & 33, 33, 33 & $\begin{pmatrix} (1.5, 0) \\ (-1.5,0) \\ (0, \frac{3\sqrt{3}}{2}) \end{pmatrix}$ & $\begin{pmatrix} 1 & 0.25  \\ 0.25 & 1 \end{pmatrix}$ \\
        \hline
        5& 2 & 5 & 66, 66, 66, 66, 66 & $\begin{pmatrix} (3, 0) \\ (0,3) \\ (-3, 0) \\ (0, -3) \\ (0, 0) \end{pmatrix}$ & $\begin{pmatrix} 0.75 & 0  \\ 0 & 0.75 \end{pmatrix}$ \\
        \hline
       6& 3 & 4 & 66, 66, 66, 66 & $\begin{pmatrix} (1, 1, 1) \\ (1,-1,-1) \\ (-1, 1, -1) \\ (-1, -1, 1) \end{pmatrix}$ & $\begin{pmatrix} 1 & 0 & 0  \\ 0 & 1 & 0 \\ 0 & 0 & 1 \end{pmatrix}$ \\
        \hline
    \end{tabular}
    \caption{Fundamental parameters of the datasets generated.}
    \label{table:datasets}
\end{table}

\begin{figure}[H]
    \centering
    \includegraphics[width=0.75\linewidth]{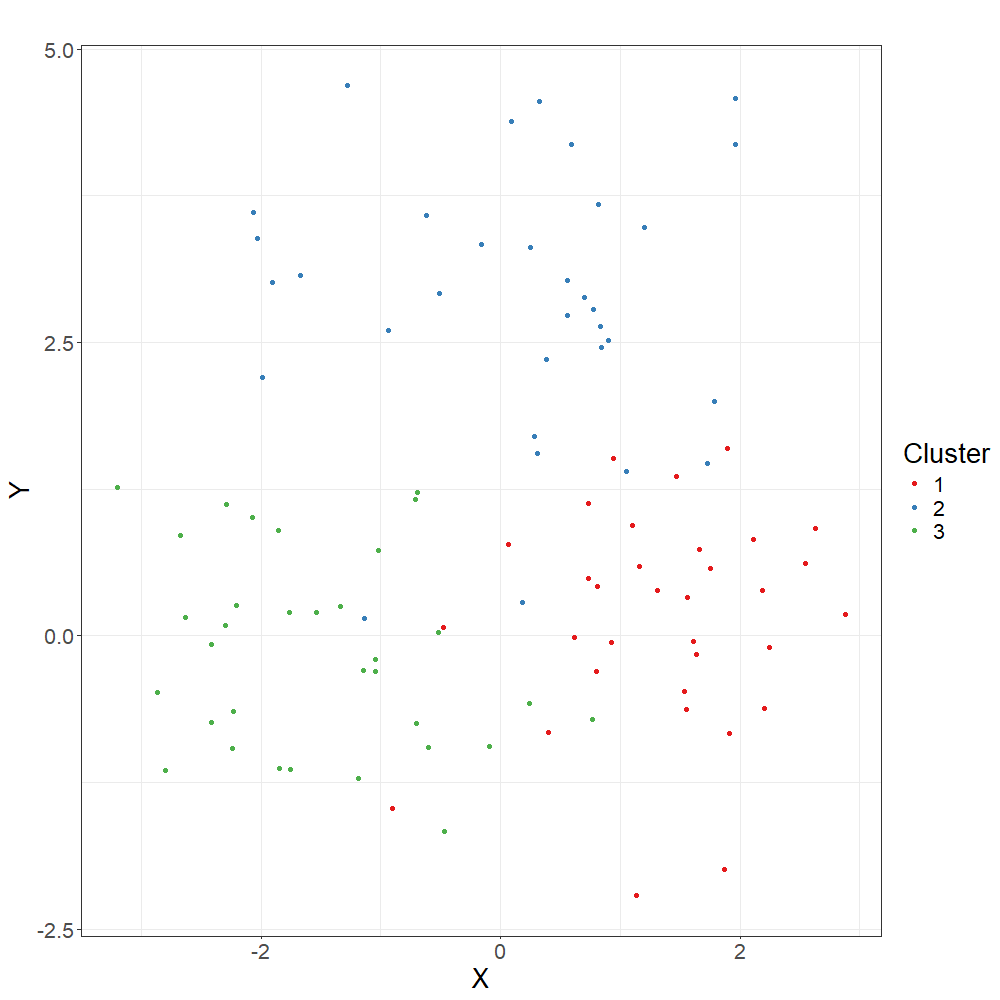}
    \caption{Visual representation of dataset 1, each colour associated with a differently generated component.}
    \label{ds1}
\end{figure}

\begin{figure}[H]
    \centering
    \includegraphics[width=0.75\linewidth]{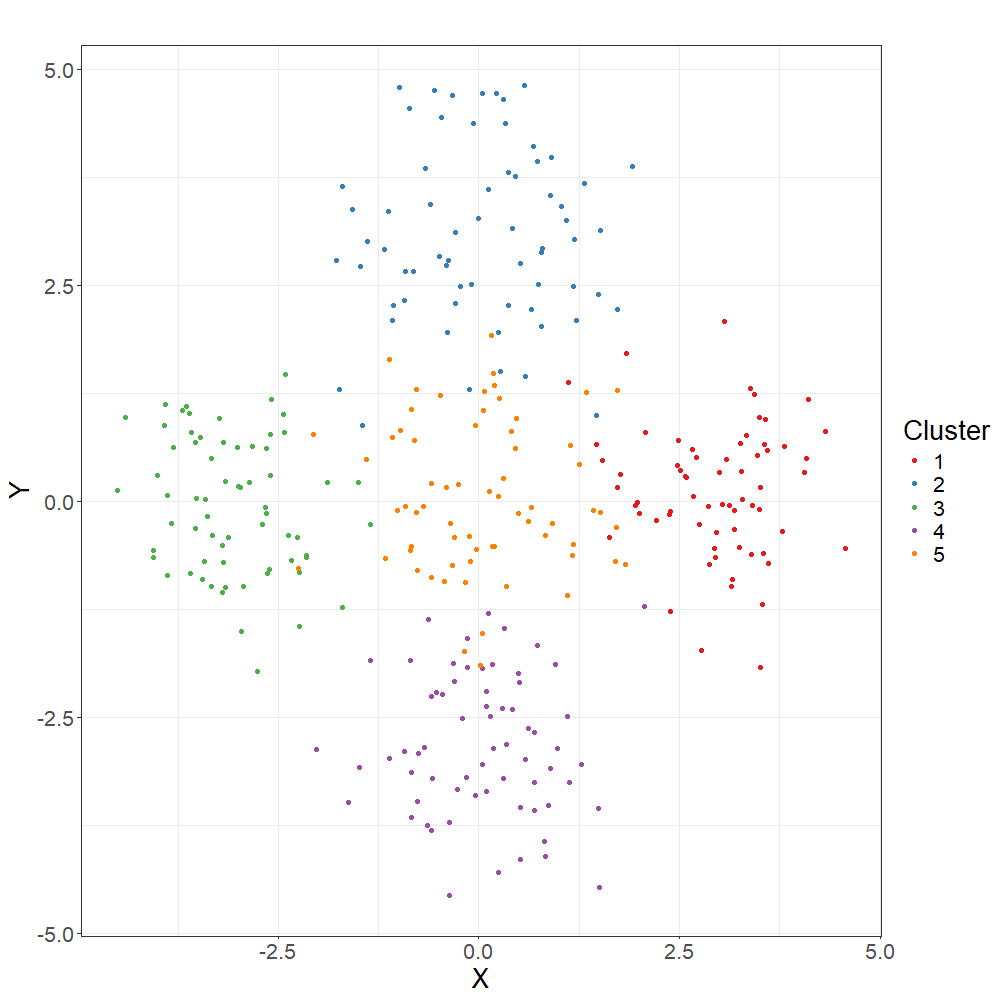}
    \caption{Visual representation of dataset 5, each colour associated with a differently generated component.}
    \label{ds5}
\end{figure}

\noindent The results obtained for each dataset are shown in Tables \ref{table:meanent} and \ref{table:maxpair}, for the measures $\Bar{S}$ and $\Bar{S}_{l,m}$ respectively. 
The method is compared against the results obtained with traditional $K$-selection methods: silhouette method \cite{silhouette}, shown in Table \ref{table:sil}, and gap statistic \cite{Elements2009}, shown in Table \ref{table:gap}.\\
Both proposed measures are able to find as optimal $K$ the one used for generation of the dataset, with the choice of parameter $s=1$; on the contrary, the silhouette method fails to do so for datasets 5 and 6, and the gap statistic for dataset 6. \\
As visual examples, datasets 1 and 5 are shown in Figure \ref{ds1} and Figure \ref{ds5}. 

\begin{table}[H]
    \centering
    \begin{tabular}{|c|c|c|c|c|c|}
        \hline
        Dataset & $K =$ 2 & 3 & 4 & 5 & 6 \\ \hline
       1& 0.120 & \textbf{0.055} & 0.139 & 0.168 & 0.207 \\ \hline
       2& 0.082 & \textbf{0.052} & 0.115 & 0.102 & 0.155 \\ \hline
       3& 0.168 & \textbf{0.113} & 0.179 & 0.177 & 0.186 \\ \hline
       4& 0.216 & \textbf{0.061} & 0.183 & 0.112 & 0.179 \\ \hline
       5& 0.036 & 0.183 & 0.049 & \textbf{0.032} & 0.124 \\ \hline
       6& 0.235 & 0.228 & \textbf{0.113} & 0.130 & 0.199
 \\ \hline
    \end{tabular}
    \caption{$\Bar{S}$, entropy $S$ averaged over the dataset, choosing different values of $K$. Results are shown for the case with parameter $s=1$. Each row refers to the corresponding dataset in Table \ref{table:datasets}. The measure takes the minima for the value of $K$ used for dataset generation, in all cases.}
    \label{table:meanent}
\end{table}


\begin{table}[H]
    \centering
    \begin{tabular}{|c|c|c|c|c|c|}
        \hline
        Dataset &$K =$ 2 & 3 & 4 & 5 & 6 \\ \hline
        1&0.173 & \textbf{0.028} & 0.091 & 0.055 & 0.067 \\ \hline
       2& 0.118 & \textbf{0.031} & 0.086 & 0.042 & 0.048 \\ \hline
        3&0.243 & \textbf{0.065} & 0.086 & 0.111 & 0.081 \\ \hline
       4& 0.311 & \textbf{0.040} & 0.148 & 0.061 & 0.076 \\ \hline
        5&0.052 & 0.134 & 0.035 & \textbf{0.020} & 0.124 \\ \hline
        6&0.340 & 0.200 & \textbf{0.049} & 0.056 & 0.079 \\ \hline
    \end{tabular}
    \caption{$\Bar{S}_{l,m}$: maximum value with respect to the cluster pairs of the pairwise entropy $S_{l,m}$ averaged over the dataset, choosing different values of $K$. Results are shown for the case with parameter $s=1$. Each row refers to the corresponding dataset in Table \ref{table:datasets}. The measure takes the minima for the value of $K$ used for dataset generation, in all cases.}
    \label{table:maxpair}
\end{table}

\vspace{-10ex}

\begin{table}[H]
    \centering
    \begin{tabular}{|c|c|c|c|c|c|}
        \hline
        Dataset &$K =$ 2 & 3 & 4 & 5 & 6 \\ \hline
        1&0.357 & \textbf{0.472} & 0.416 & 0.343 & 0.359 \\ \hline
        2&0.403 & \textbf{0.457} & 0.417 & 0.351 & 0.354 \\ \hline
       3& 0.337 & \textbf{0.387} & 0.347 & 0.325 & 0.335 \\ \hline
       4& 0.409 & \textbf{0.474} & 0.388 & 0.398 & 0.351 \\ \hline
       5& 0.362 & 0.392 & \textbf{0.469} & 0.458 & 0.430 \\ \hline
        6&0.321 & \textbf{0.365} & 0.352 & 0.343 & 0.348 \\ \hline
    \end{tabular}
    \caption{Value of the silhouette parameter for different values of $K$. In bold are the maximum values, indicating the optimal $K$ for the silhouette method. Each row refers to the corresponding dataset in Table \ref{table:datasets}.}
    \label{table:sil}
\end{table}

\vspace{-10ex}

\begin{table}[H]
    \centering
    \begin{tabular}{|c|c|c|c|c|c|}
        \hline
        Dataset & $K =$ 2 & 3 & 4 & 5 & 6 \\ \hline
       1& 0.135 & \textbf{0.253} & 0.159 & 0.132 & 0.149 \\ \hline
        2&0.312 & \textbf{0.338} & 0.239 & 0.256 & 0.262 \\ \hline
       3& 0.179 & \textbf{0.200} & 0.132 & 0.092 & 0.107 \\ \hline
       4& 0.173 & \textbf{0.238} & 0.171 & 0.159 & 0.113 \\ \hline
       5& 0.132 & 0.177 & 0.261 & \textbf{0.272} & 0.247 \\ \hline
      6& \textbf{0.278} & 0.267 & 0.209 & 0.223 & 0.219 \\ \hline
    \end{tabular}
    \caption{Value of the gap statistic for different values of $K$. In bold are the maximum values, indicating the optimal $K$ for the gap statistic method. Each row refers to the corresponding dataset in Table \ref{table:datasets}.}
    \label{table:gap}
\end{table}

\begin{figure}[H]
    \centering
    \includegraphics[width=0.8\linewidth]{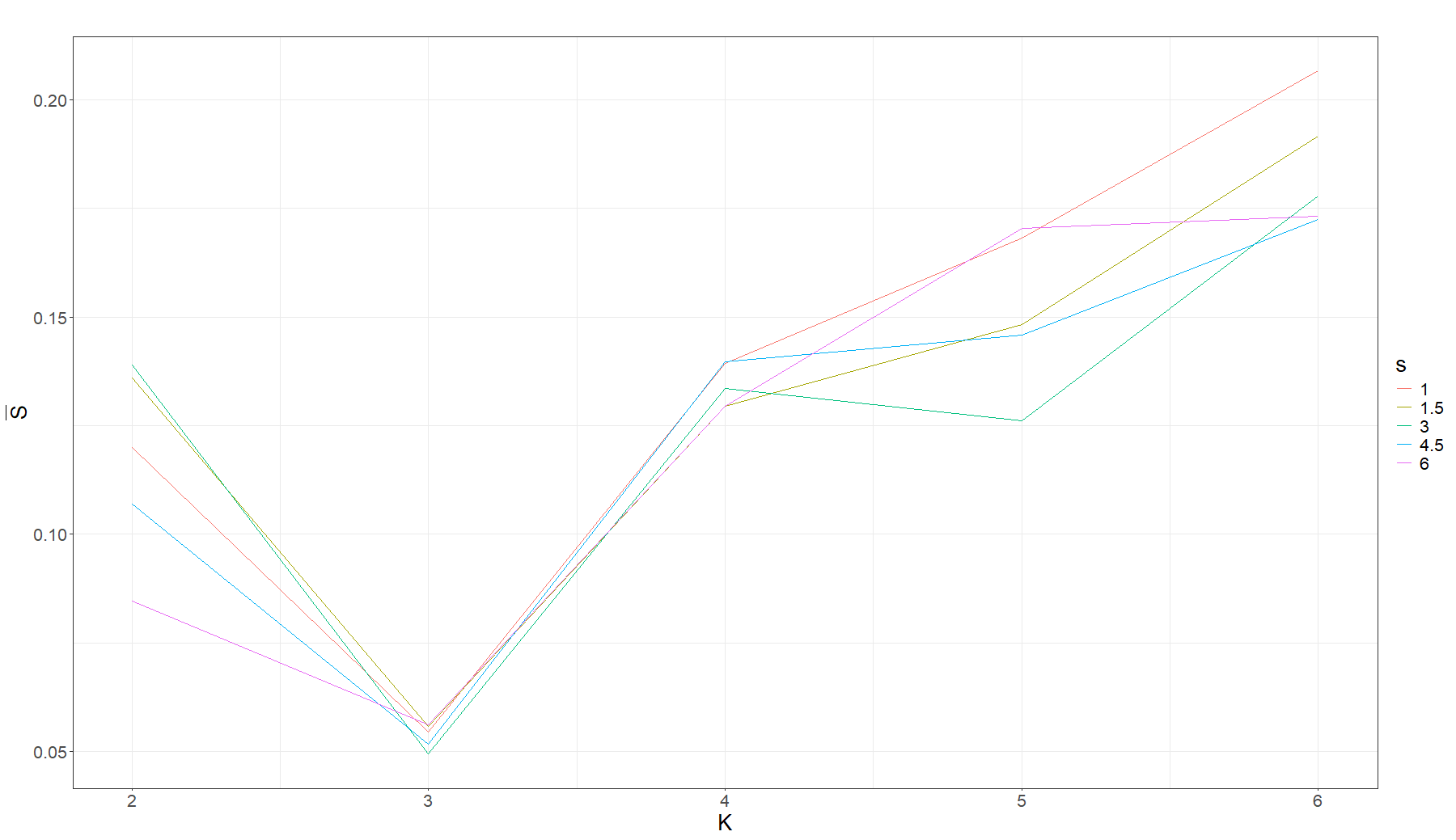}
    \caption{Dataset 1, behaviour of the measure $\Bar{S}$, averaged over the dataset, as function of the chosen number of clusters $K$. Each line represents the results of the method with different values of the parameter $s= 1, 1.5, 3, 4.5, 6$. The measure takes the minimum for $K = 3$, the number of clusters used in the dataset generation.}
    \label{fig:m1_}
\end{figure}

\begin{figure}[H]
    \centering
    \includegraphics[width=0.8\linewidth]{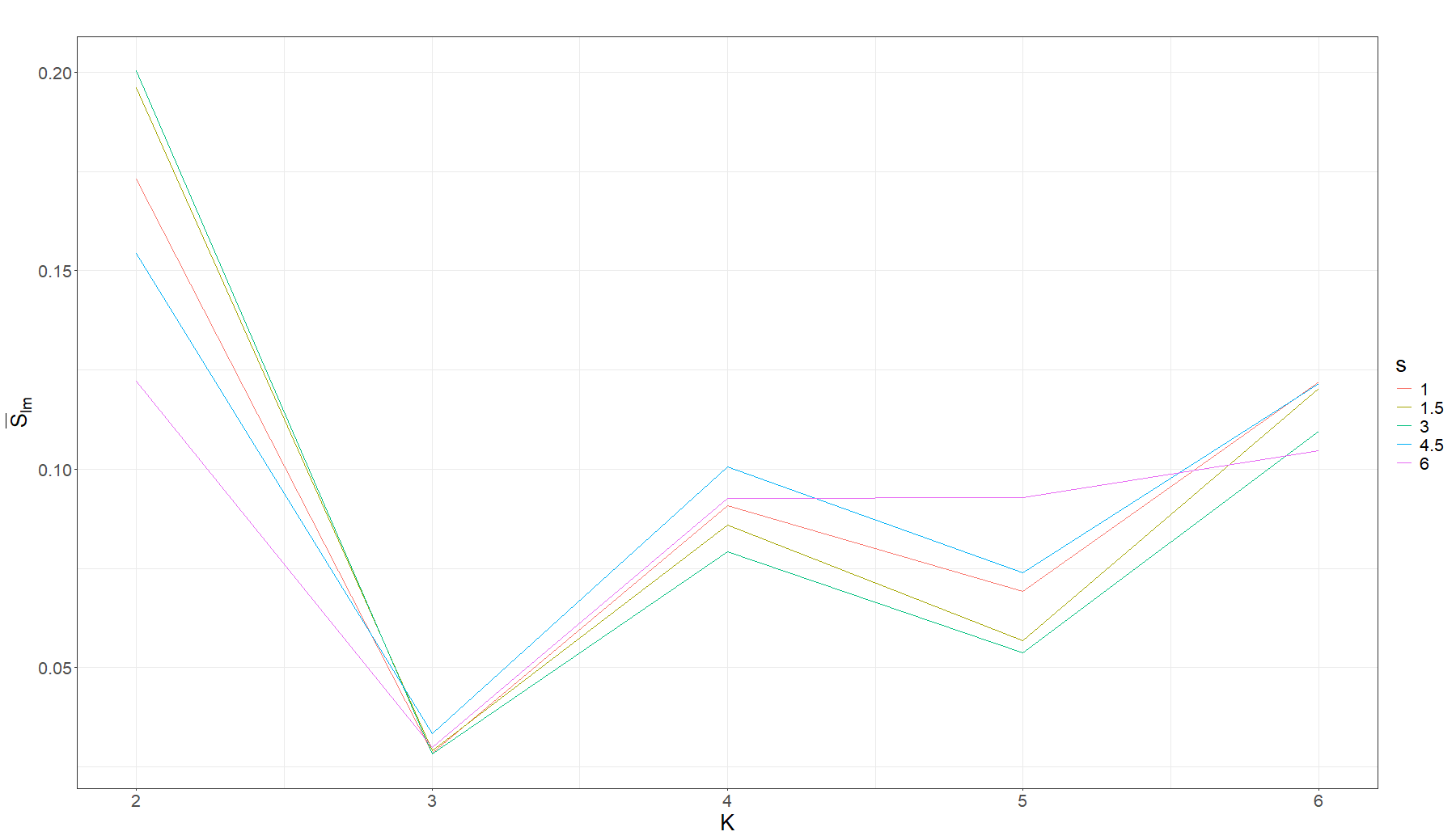}
    \caption{Dataset 1, behaviour of the measure $\bar{S}_{l,m}$ as function of the chosen number of clusters $K$. Each line represents the results of the method with different values of the parameter $s= 1, 1.5, 3, 4.5, 6$. The measure takes the minimum for $K = 3$, the number of clusters used in the dataset generation.}
    \label{fig:m2_}
\end{figure}

\noindent Regarding dataset 1, in Figure \ref{fig:m1_} and Figure \ref{fig:m2_} are shown the values of the two measures for varying $K$, with different choices of parameter $s$: the measures tend to be robust with respect to the different choices of informativeness parameter $s$. In fact, all the curves show a minimum for K=3. \\

\vspace{-6ex}

\begin{figure}[H]
    \centering
    \includegraphics[width=0.8\linewidth]{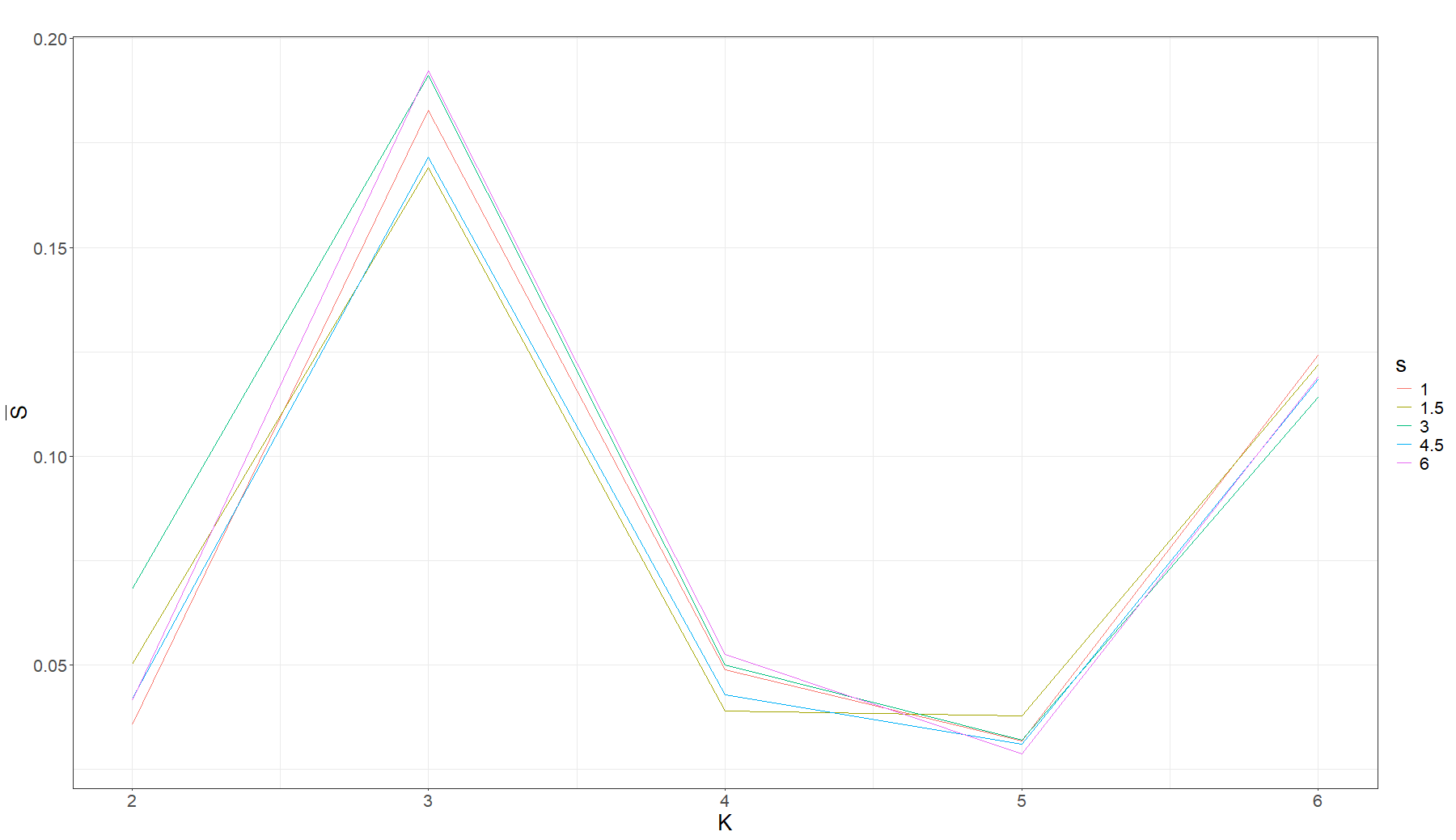}
    \caption{Dataset 5, behaviour of the measure $\bar{S}$, averaged over the dataset, as function of the chosen number of clusters $K$. Each line represents the results of the method with different values of the parameter $s= 1, 1.5, 3, 4.5, 6$. The measure takes the minimum for $K = 5$,  the number of clusters used in the dataset generation, in all cases except when $s=1.5$.}
    \label{fig:m1_5}
\end{figure}

\begin{figure}[H]
    \centering
    \includegraphics[width=0.8\linewidth]{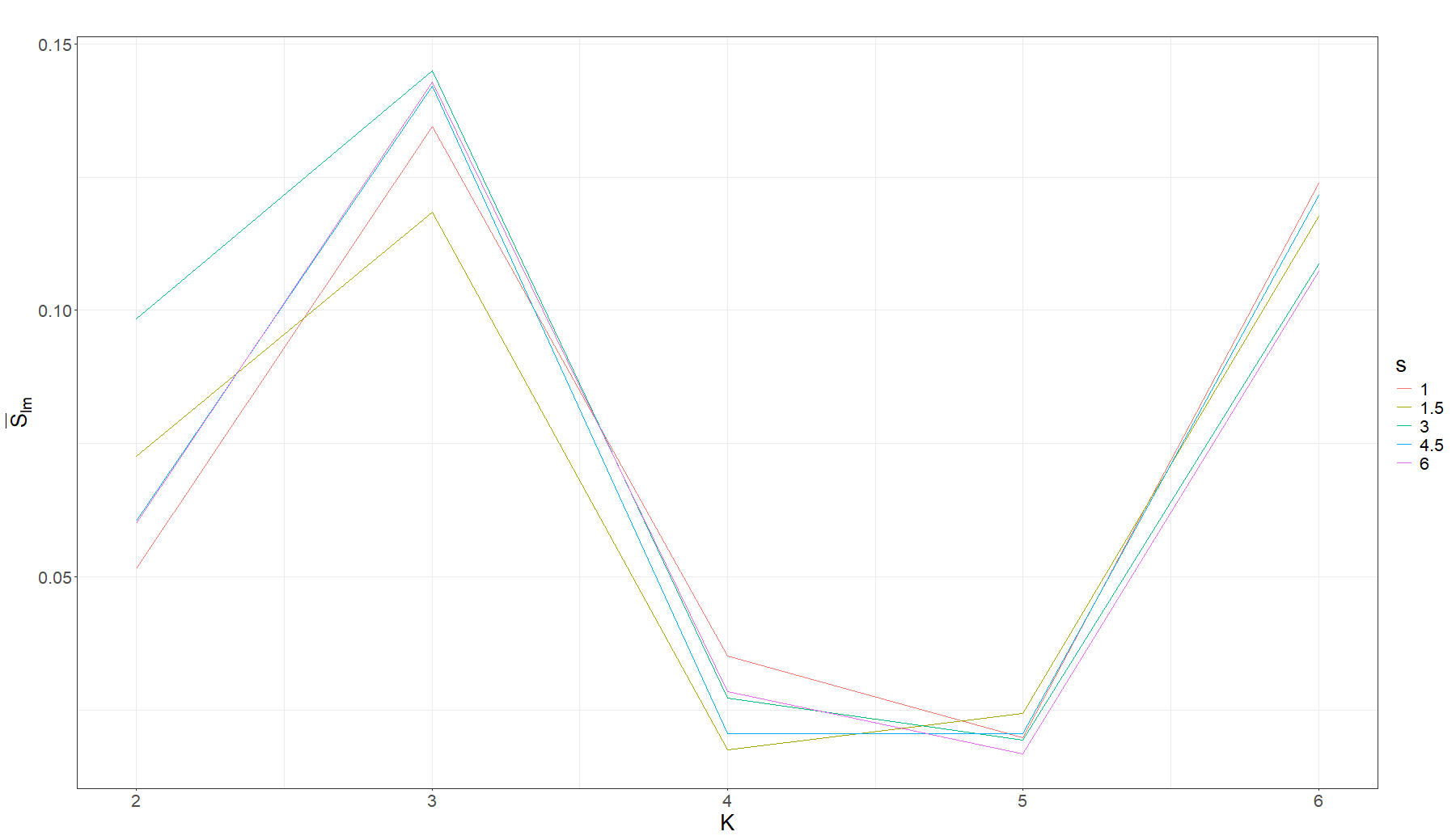}
    \caption{Dataset 5, behaviour of the measure $\bar{S}_{l,m}$ as function of the chosen number of clusters $K$. Each line represents the results of the method with different values of the parameter $s= 1, 1.5, 3, 4.5, 6$. The measure takes the minima for $K = 4$ or $K = 5$, depending on the parameter $s$.}
    \label{fig:m2_5}
\end{figure}

\noindent The values of the two measures for varying $K$, with different choices of parameter $s$, are shown for dataset 5 in Figures \ref{fig:m1_5} and \ref{fig:m2_5}. The difficulty in choosing between $K=4,5$ is highlighted by the fact that different values of $s$ correspond to different optima, especially in Figure \ref{fig:m2_5}. This behaviour means that the general uncertainty of assignments is minimum for $K=5$, even if the difference is not so big with $K=4$, but going from 4 to 5 clusters does not reduce the uncertainty between a pair of clusters.\\

\noindent We underline that the method can be used with two different perspectives. As a wrapped method for a quick choice of K or as a tool to explore in a deeper way the geometric structure of data, led by the idea of entropy of the memberships.

\newpage
\section{Conclusions}
In this work we proposed a new clustering approach, based on the deployment of bagging techniques, which enhances a large family of existing methods. With little computational effort, the method adds the benefit of a Bayesian interpretation of the data generating process, 
at the same time having the potential to retrieve additional information of participation while preserving (and in some cases bettering) the benchmark of usual methods. 
The proposal of an optimal scheme for choosing the cluster number follows naturally from the characteristics of the clustering proposal: the introduction of prior knowledge in the clustering procedure enforces regular $K$-cluster behaviours, while the fuzzy information retrieved leads to quantifiable measures aimed at evaluating optimality.
Future directions of work include the usage of different methods from K-means, 
more refined choices of prior, and the extension of the method to other types of aggregation used in a cluster setting (e.g. BagClust2, \cite{dudoit2003bagging}).
Further analysis will be also carried out for different datasets, focusing particularly on dimensionality effects. \\

\section*{Acknowledgements}
F.M. Quetti acknowledges RES, in particular C.E.O. Federico Bonelli for the award of Ph.D. scholarship.

\end{document}